\documentclass[letterpaper, 10 pt, conference]{ieeeconf}
\IEEEoverridecommandlockouts                              

\usepackage{cite}
\usepackage{multirow}
\usepackage{algorithm}  	
\usepackage{algpseudocode}
\usepackage{amsmath}
\usepackage{graphics}
\usepackage{epsfig}

\usepackage{graphicx}
\usepackage{times}
\usepackage{amsmath}
\usepackage{amssymb}
\usepackage{graphics} 
\usepackage[utf8]{inputenc}
\usepackage{boldline}
\usepackage{multicol}

\usepackage{url}            
\usepackage{booktabs}       
\usepackage{amsfonts}       
\usepackage{nicefrac}       
\usepackage{microtype}      
\usepackage{bm}
\usepackage{epsfig}
\usepackage{diagbox}
\usepackage{makecell}
\usepackage{colortbl}
\usepackage{threeparttable}
\usepackage{wrapfig}
\usepackage[dvipsnames]{xcolor}
\usepackage{epstopdf}
\usepackage[colorinlistoftodos]{todonotes}
\usepackage[capitalise]{cleveref}
\usepackage{algpseudocode}
\usepackage{amsmath}
\usepackage{subfigure}

\usepackage{amsmath,amsfonts,bm}

\def\eqref#1{equation~\ref{#1}}

\def\1{\bm{1}}

\def\vf{{\bm{f}}}
\def\vg{{\bm{g}}}

\def\vv{{\bm{v}}}

\def\vx{{\bm{x}}}

\def\mC{{\bm{C}}}

\def\mF{{\bm{F}}}

\def\mI{{\bm{I}}}

\def\mP{{\bm{P}}}

\def\mR{{\bm{R}}}
\def\mS{{\bm{S}}}

\def\mW{{\bm{W}}}
\def\mX{{\bm{X}}}

\def\mZ{{\bm{Z}}}

\DeclareMathAlphabet{\mathsfit}{\encodingdefault}{\sfdefault}{m}{sl}
\SetMathAlphabet{\mathsfit}{bold}{\encodingdefault}{\sfdefault}{bx}{n}

\def\sG{{\mathbb{G}}}

\def\sP{{\mathbb{P}}}

\def\sS{{\mathbb{S}}}

\newcommand{\R}{\mathbb{R}}

\title{\LARGE \bf
Elastic Interaction of Particles for Robotic Tactile Simulation
}

\author{Yikai Wang, Wenbing Huang, Bin Fang, Fuchun Sun$^\dagger$\thanks{$^\dagger$ Fuchun Sun is the corresponding author.}
\\Beijing National Research Center for Information Science and Technology$\,$(BNRist),\\ State Key Lab on Intelligent Technology and Systems,\\ Department of Computer Science and Technology, Tsinghua University \\ \texttt{\footnotesize wangyk17@mails.tsinghua.edu.cn, hwenbing@126.com, \{fangbin, fcsun\}@tsinghua.edu.cn}}
\begin{document}

\maketitle
\thispagestyle{empty}
\pagestyle{empty}

\begin{abstract}
Tactile sensing plays an important role in robotic perception and manipulation. To overcome the real-world limitations of data collection, simulating tactile response in virtual environment comes as a desire direction of robotic research. Most existing works model the tactile sensor as a rigid multi-body, which is incapable of reflecting the elastic property of the tactile sensor as well as characterizing the fine-grained physical interaction between two objects. In this paper, we propose Elastic Interaction of Particles (EIP), a novel framework for tactile emulation. At its core, EIP models the tactile sensor as a group of coordinated particles, and the elastic theory is applied to regulate the deformation of particles during the contact process. The implementation of EIP is conducted from scratch, without resorting to any existing physics engine. 
Experiments to verify the effectiveness of our method have been carried out on two applications: robotic perception with tactile data and 3D geometric reconstruction by tactile-visual fusion.
It is possible to open up a new vein for robotic tactile simulation, and contribute to various downstream robotic tasks. 
\end{abstract}

\section{Introduction}
Tactile sensing is one of the most compelling perception pathways for nowadays robotic manipulation, as it is able to deliver, from the contact between sensor and object, the physical patterns including shape, texture, and physical dynamics that are not easy to perceive via other modalities, \emph{e.g.} vision. In recent years, data-driven machine learning approaches have exploited tactile data and exhibited the success in a variety of robotic tasks, such as object recognition~\cite{liu2017recent}, grasp stability detection~\cite{kwiatkowski2017grasp,zapata2019tactile}, and manipulation~\cite{fang2018dual,tian2019manipulation} to name some. 

That being said, the learning-based methods---particularly those involving deep learning---are usually data-hungry and require large datasets for model training. 
Collecting large real tactile dataset is not easy to fulfil since it demands continuous robot control which is time-consuming or even risky considering the hardware wear and tear. Another concern with real tactile collection is that the data acquired by the sensors of different shape/material or under different control policy could be heterogeneously distributed, posing a challenge to fairly assess the effectiveness of different learning methods trained on different tactile datasets.  

The simulation of the tactile sensing can potentially help overcome these real-world limitations. Via simulation, one can easily build up a large dataset for tactile benchmarking without the time or source constraint. Yet, establishing a promising tactile simulator, by no means, is challenging. We need to know how to geometrically and physically model the tactile senor, and more importantly, we should be capable of characterizing the physical interaction during the contact process between the sensor and the object, both of which make simulating tactile data more difficult than other modalities, such as vision that is solely geometrically-aware.

\begin{figure}[t]
\centering
\includegraphics[scale=0.44]{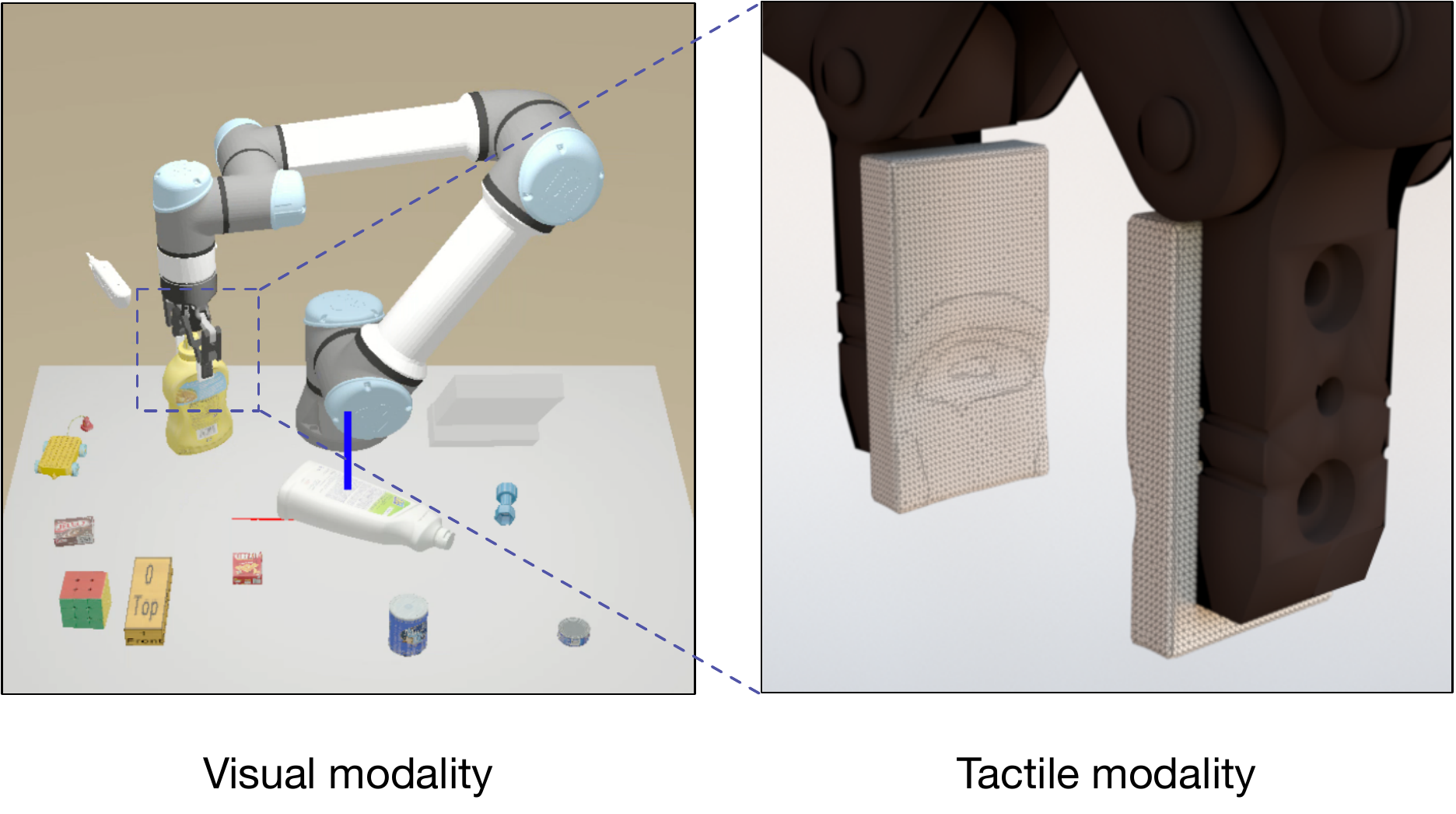}
\caption{Illustration of the simulated tactile pattern when a robot is picking up a mustard bottle. Our framework enables researchers to deal with tactile-visual multimodal information in the simulated environment.}
\label{channel}
\end{figure}

There have been several trails that consider to simulate tactile interactions with manipulated objects~\cite{moisio2013model,kappassov2020simulation,sferrazza2020learning,ding2020sim}. Most existing methods model the tactile sensor as a combination of rigid bodies, and the collision between two objects is described by rigid multi-body kinematics provided by certain off-the-shelf physics
engine (such as ODE in~\cite{kappassov2020simulation}). Despite its validity in some cases, considering the tactile sensor as a rigid multi-body will overlook the fact that common tactile sensors are usually elastic but not rigid. For example, the sensors invented by~\cite{yuan2017gelsight} leverages elastic materials to record the deformation to output tactile sensing. Moreover, in current methods, the segmentation of tactile sensor into rigid bodies is usually coarse and the interaction between rigid bodies is unable to capture the high-resolution sensor-object contact.

In this paper, we propose a novel methodology for tactile simulation, dubbed as Elastic Interaction of Particles (EIP). EIP first models the tactile sensor as a group of coordinated particles of certain mass and size. By assuming the sensor to be made of elastic materials, the elastic theory is applied to constraint the movement of particles~\cite{stomakhin2012energetically}. During the interaction between the sensor and the object, the deformation of particles are recorded as the tactile data, which is updated via the method proposed by~\cite{hu2018moving}.

To sum up, our contributions are as follows.
\begin{itemize}
    \item We propose EIP, a novel tactile simulating framework that is capable of modeling the elastic property of the tactile sensor and the fine-grained physical interaction between the sensor and the object. 
    \item In contrast to existing methods that usually exploit the off-the-shelf physics engine for interaction simulation, the implementation of our method is formulated from scratch, which makes our framework more self-contained and easier to be plugged into down-stream applications. 
    \item We evaluate the effectiveness of our method on two applications: robotic perception with tactile  data and  3D geometric  reconstruction by  tactile-visual fusion. Both applications well support the benefits of our proposed idea, as demonstrated by our experiments.
\end{itemize}

\vspace{0.1 in}
\section{Related Work}
Nowadays the vision based tactile sensors have become prominent due to their superior performance of robotic perception and manipulation. Data-driven approaches to tactile sensing are commonly used to overcome the complexity of accurately modeling contact with soft materials. However, their widespread adoption is
impaired by concerns about data efficiency and the capability to generalize when applied to various tasks.
Hence simulation approaches of vision-based tactile sensing are developed recently. 

Regarding the exploration of tactile simulation, early work \cite{journals/trob/ZhangC00} that directly adopts the elastic theory for the mesh interaction resorts to high computation costs. 
\cite{moisio2013model} represents the tactile sensor as a rigid body, and calculates the interaction force on each triangle mesh.  
\cite{habib2014skinsim} models the tactile sensor as rigid elements, and simulates their displacement by adding a virtual spring, with help of the commonly used Gazebo simulator. Modeling the tactile sensor as one or a combination of independent rigid bodies makes these methods difficult to obtain high-resolution tactile patterns, and these methods also overlook the fact that real tactile sensors are mostly elastic materials. 
\cite{ding2020sim} designs a model for soft body simulation which is implemented using the Unity physics engine, and trains a neural network to predict the locations and angles of edges when contacting with the sensor.  \cite{gomes2019gelsight} introduces the approach for simulating a GelSight tactile sensor in the Gazebo simulator, by directly modelling the contact surface without considering the material of the tactile sensor. Based on costly Finite-Element Analysis, \cite{sferrazza2020learning} provides the simulation strategy to generate an entire supervised learning dataset for a vision-based tactile sensor, with the objective of estimating the full contact force distribution from real-world tactile images. 

\begin{figure*}[t]
\centering
\includegraphics[scale=0.53]{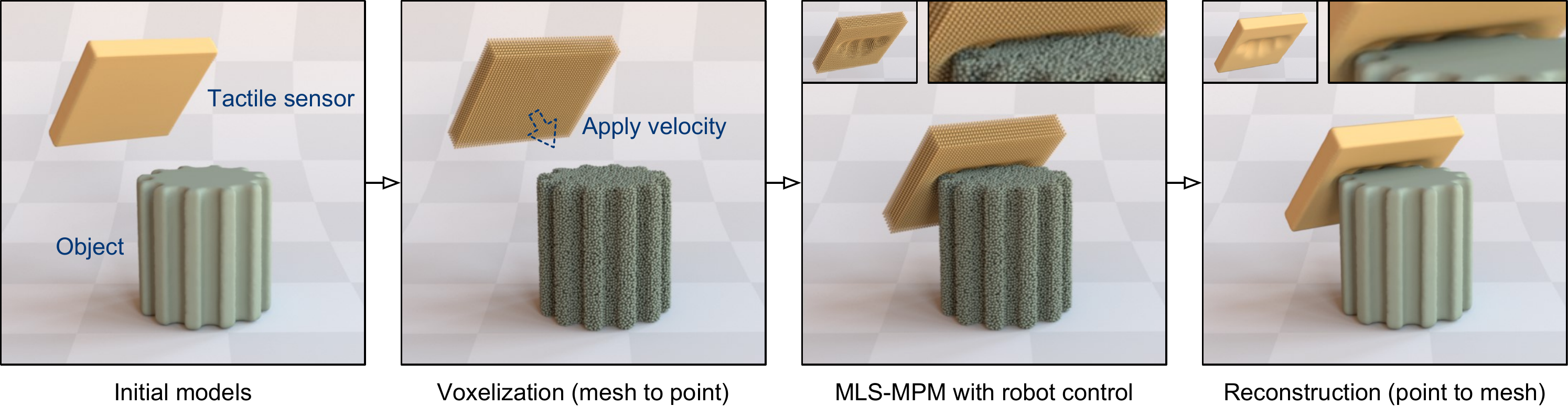}
\caption{A brief framework of our tactile simulation process. The initial 3D meshes are first converted to particles by voxelization. The particles of tactile sensor and the manipulated object are interacted by MLS-MPM, with additional control at the robot side. Meshes can be optionally  reconstructed from the points for rendering purpose.}
\label{fig:channel}
\end{figure*}

\section{Physically Tactile Simulation}

In this section, we first introduce how to simulate the physical interaction between the tactile sensor and the targeted object. 
We then evaluate the effectiveness of our simulated tactile sensor via two applications: robotic perception with tactile sensing and 3D geometric reconstruction by tactile-visual fusion. 

\subsection{Overall Framework}

The basic idea of our method is to assume both the sensor, and the object to be solid and model them in the form of particles. In general, the tactile simulation process, as depicted in Figure~\ref{fig:channel}, consists of 3 steps: voxelization from meshes to particles, interaction simulation, and reconstruction from particles to meshes. We detail each step below.

\paragraph{Voxelization}
We obtain from the simulation environment the triangle meshes that describe the geometric model of the object/sensor. The inside of each model is filled with dense voxel grids by using the method of voxel carving. Briefly, we first calculate the depth maps and then employ these depth maps to carve a dense voxel grid. We refer readers to \cite{Zhou2018} for more details. The center of each voxel grid is denoted as a particle.  

\paragraph{Interaction Simulation}
After voxelization, we apply a certain velocity to the tactile sensor until it touches the object to a certain extent. How the particles of the tactile sensor deform implies what the interaction process is. We simulate the deformation process based on Material-Point-Method (MPM)~\cite{stomakhin2013material} and its modification MLS-MPM~\cite{hu2018moving} considering both efficacy and efficiency, where we also consider the specific movement of the tactile sensor under robot control. The details of this step is provided in \textsection~\ref{sec:interaction}.  

\paragraph{Reconstruction}
The final step is to reconstruct the meshes based on the positions of particles, which can be accomplished by using the method proposed by~\cite{kazhdan2013screened}. Note that this step is not necessary unless we want to render the interaction at each time step.

\subsection{Interaction Simulation}
\label{sec:interaction}
This subsection presents the details of how we simulate the interaction process. 
In practice, the tactile sensor is basically made of elastic materials, and our main focus is on the change of its shape, or called deformation. We apply the elastic theory to constrain the deformation of the particles in the tactile sensor during the interaction with the object, and the deformation at each time step will be recorded as the tactile sensing data.

Suppose that the sensor is composed of $m$ particles. The coordinate of the $p$-th particle is represented by $\vx_p\in\R^{d}$, where $d=3$ throughout our paper. We define the deformation map as $\bm{\Phi}: \R^{d}\rightarrow \R^{d}$. The Jacobian of $\bm{\Phi}$ with respect to the $p$-th particle, denoted as $\mF_p\in\R^{d\times d}$ (\emph{a.k.a} deformation gradient), is calculated by
\begin{eqnarray}
\label{eq:deformation-gradient}
\mF_p = \frac{\partial \bm{\Phi}}{\partial \vx}(\vx_p).
\end{eqnarray}
When the particle deforms, its volume may also change. The volume ratio by the deformation, denoted as $J_p$, is the determinant of $\mF_p$, \emph{i.e.},
\begin{eqnarray}
J_p=\text{det}(\mF_p).
\end{eqnarray}

To describe the stress–strain relationship for elastic materials, a strain energy density function $\bm{\Psi}$ is adopted, which is a kind of potential function that constrains deformation $\mF_p$. We follow a widely used method called Fixed Corotated~\cite{stomakhin2012energetically}, which computes $\bm{\Psi}$ by
\begin{eqnarray}
\label{eq:phi}
    \bm{\Psi}(\mF_p) =\mu\sum_{i=1}^d(\sigma_{i,p}-1)^2+\frac{\lambda}{2}(J_p-1)^2,
\end{eqnarray}
where $\mu=\frac{E}{2(1+v)}$ and $\lambda=\frac{Ev}{(1+v)(1-2v)}$ are Lamé's 1st and 2nd parameters, respectively, and $E,v$ are Young's modulus and Poisson ratio of the elastic material, respectively; $\sigma_{i,p}$ is the $i$-th singular value of $\mF_p$. The derivative of $\bm{\Psi}$ (\emph{a.k.a} the first Piola-Kirchhoff stress) will be utilized to adjust the deformation process, which can be derived by
\begin{eqnarray}
\label{eq:PK}
    \mP_p=\frac{\partial \bm{\Psi}}{\partial \mF}(\mF_p)=2\mu(\mF_p-\mR_p)+\lambda(J_p-1)J_p\mF_p^{-\mathsf{T}},
\end{eqnarray}
where $\mR_p$ is obtained via the polar decomposition~\cite{higham1986computing}: $\mF_p=\mR_p\mS_p$.

In the following context, we will characterize how each particle deforms, that is, how its position $\vx_p$ changes, during the interaction phase. For better readability, we distinguish the position $\vx_p$ and the quantities in Eq.~(\ref{eq:deformation-gradient}-\ref{eq:PK}) at different time step by adding a temporal superscript, \emph{e.g.} denoting the velocity at time step $n$ as $\vx_p^{(n)}$. We leverage MLS-MPM to update $\vx_p^{(n)}$. The core of this method is dividing the whole space into grids of certain size. For each particle, its velocity is updated as the accumulated velocity of all particles within the same grid, which, to some extend, can emulate the physical interaction between particles. 
Specifically, we iterate the following steps for the update of $\vx_p^{(n)}$. The flowchart is sketched in Algorithm~\ref{alg:tactile_simulation}.

\paragraph{Momentum Scattering}
For each grid, we collect the mass and the momentum from the particles inside and those within its neighbors. The mass of the $i$-th grid is collected by 
\begin{equation}
\label{eq:mass}
    m'_i = \sum_{j\in\sG_i}\sum_{p\in\sP_j} w_{jp}m_p,
\end{equation}
where, $\sG_i$ denotes the $3\times3\times3$ grids surrounding grid $i$, $\sP_j$ collects the indices of the particles located in grid $j$, $m_p$ denotes the mass of the $p$-th particle, and $w_{jp}$ computes the B-Spline kernel negatively related to the distance between the $j$-th grid and the $p$-th particle. 

The momentum of the $i$-th grid is derived as
\begin{align}
\label{eq:momentum}
(m'\vv')_i &=\sum_{j\in\sG_i}\sum_{p\in\sP_j} w_{jp} \left(m_p\vv_p^{(n)}+\mC_p^{(n)}(\vx'_j-\vx_p^{(n)})\right)\notag\\
&-\sum_{j\in\sG_i}\sum_{p\in\sP_j} w_{jp}\gamma\mP_p^{(n)}(\mF_p^{(n)})^{\mathsf{T}}(\vx'_j-\vx_{p}^{(n)}),
\end{align}
where $\vx'_j$ denotes the position of grid $j$, $v_p^{(n)}$ is the velocity of particle $p$,  $\mC_p^{(n)}\in\R^{d\times d}$, adopted as an approximated parameter in~\cite{hu2018moving}, is associated to the particle $p$ whose update will be specified later, and $\gamma=\frac{4\Delta t}{\Delta x'^2} V_p^0$ is a fixed coefficient where $\Delta t$ is the time interval, $\Delta x'$ is the spatial interval between grids, and $V_p^0$ is the initial particle volume. 

\paragraph{Velocity Alignment}
The velocity on the $i$-th grid can be obtained given the grid momentum and the grid mass by normalization, 
\begin{eqnarray}
\label{eq:grid_velocity}
    \vv'_i=\frac{(m'\vv')_i}{m'_i}.
\end{eqnarray}
Note that the grid velocity is only for later parameter updating, and the position of the grid will not change in the simulation. 

\begin{figure}[t]
\centering
\includegraphics[scale=0.42]{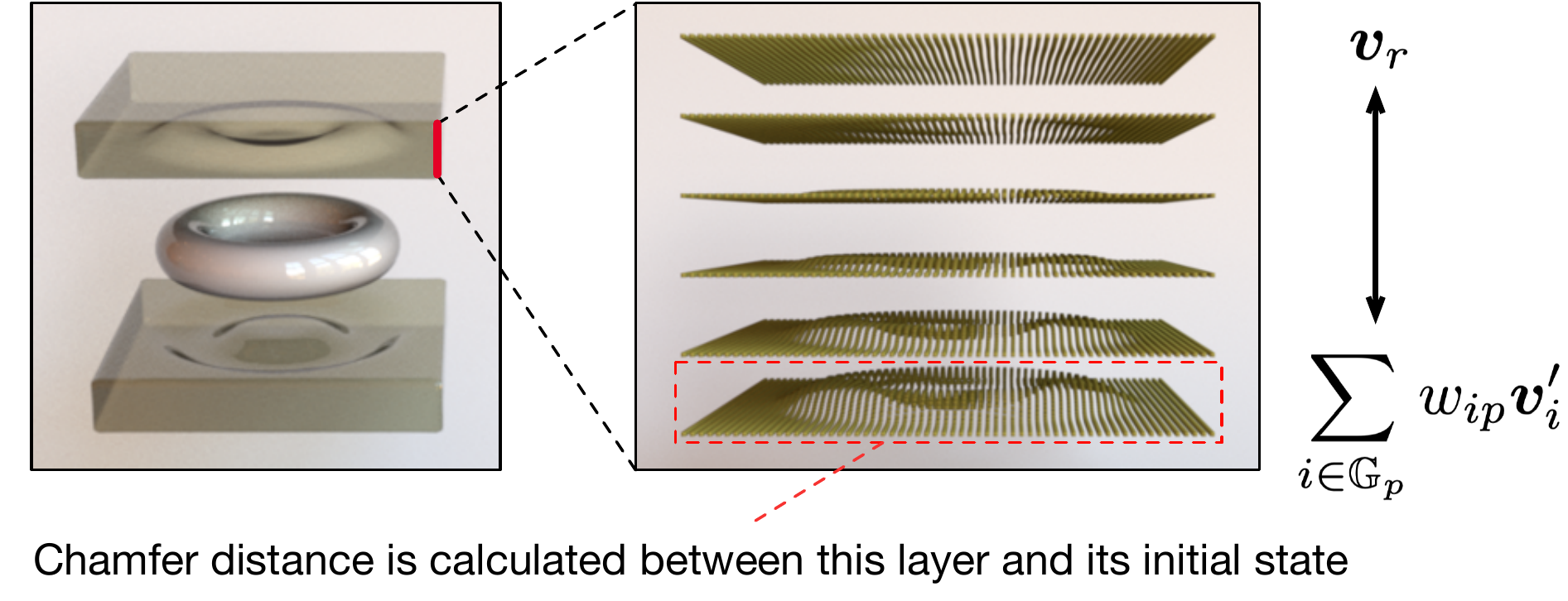}
\caption{Deformation of pressing a torus mesh.  To describe the transition from velocity of the robot hand and the particle interacted velocity, we depict the deformation of each layer. The layer that directly contacts with the manipulated object has the largest extent of deformation, and we also use this layer to calculate the chamfer distance for terminal checking. Note that we increase the distance of two tactile sensors for a better visualization.}
\label{pic:chamfer}
\vskip -0.1 in
\end{figure}

\paragraph{Robot Hand Movement}
Apart from the displacement caused by physical interaction, the sensor will also change its position due to the movement by robot hand where the sensor is equipped. This additional velocity is a local parameter depending on the position of a particle, and we use a weight $\alpha_p$ to reflect it during the scattering process while keeping its sufficient physical interaction at the contact side. The velocity is thus updated as 
\begin{align}
\label{eq:particle_v}
\vv_p^{(n+1)}&=\alpha_p\sum_{i\in\sG_p} w_{ip}\vv'_i+(1-\alpha_p)\vv_r,
\end{align}
where, $\sG_p$ is the set of $3\times3\times3$ grids where the particle $p$ is adhere to, $\vv_r$ is the velocity of the robot hand, and $\alpha_p$ is proportional to the distance between the particle and the robot hand as illustrated in Figure~\ref{pic:chamfer}.

\paragraph{Object Constraint Detection}
We assume the target object is a rigid body. Under this assumption, the velocity of the particle of the sensor will decrease to zero once it reaches the region of the object. Note that our above process is also applicable for the case when the object is supposed to be soft, which will be left for future exploration.

\paragraph{Parameters Gathering}
With the updated velocity, we renew the values of the velocity gradient, position vector, and deformation gradient by
\begin{align}
\label{eq:particle_c}
\mC_p^{(n+1)}&=\frac{4}{\Delta x^2}\sum_{i\in\sG_p} w_{ip}\vv_p^{(n+1)}(\vx_i-\vx_p^{(n)}),\\
\label{eq:particle_x}
\vx_p^{(n+1)}&=\vx_p^{(n)} + \Delta t \vv_p^{(n+1)},\\
\label{eq:particle_f}
\mF_p^{(n+1)}&=(\mI + \Delta t \mC_p^{(n+1)})\mF_p^{(n)}.
\end{align}



\begin{algorithm}[t]
	\small
	\caption{\small{Elastic Interaction of Particles (EIP)}}
	\begin{algorithmic}[1]
		\Require 3D meshes of the manipulated object and the tactile sensor, and robot hand movement velocity $\vv_{r}$; The values of Lamé's parameters: $E$ and $v$, and the massive of each particle $m_p$.
		\Ensure Tactile interaction between the object and the sensor.
		\State Convert meshes to particles using voxelization.
		\State Initialize the values of $\vx_p^{(0)}$, $\mF_p^{(0)}$, $\mC_p^{(0)}$, and $\vv_p^{(0)}$. 
		\State Dividing the whole space into grids.
		\While{not terminal}
		\For{each grid $i$}
		\State Scatter the mass and momentum of grid $i$ by Eq.~(\ref{eq:mass}-\ref{eq:momentum})
		\State Update the grid velocity by Eq.~(\ref{eq:grid_velocity})
		\EndFor
		\For{each particle $p$}
		\State Gather the velocity $\vv_p^{(n)}$ by Eq.~(\ref{eq:particle_v})
		\State Update parameters $\vx_p^{(n)}$, $\mF_p^{(n)}$, and $\mC_p^{(n)}$ by Eq.~(\ref{eq:particle_c}-\ref{eq:particle_f})
		\EndFor
		\State  Terminal check of the robot control by Eq.~(\ref{eq:chamfer})
		\EndWhile
		\State Reconstruct from particles to  meshes by Screened particle positions $\vx_{p}^{(n)}$.

	\end{algorithmic}
	\label{alg:tactile_simulation}

\end{algorithm}

\paragraph{Terminal Checking}
For safety and keeping consistent with practical usage, we will terminate the robot hand movement once the deformation of the sensor is out of certain scope. For this purpose, we use the chamfer distance to measure the distance between the deformed state and the original state of the particles in the contact surface. In form, we compute
\begin{equation}
\label{eq:chamfer}
   l=\sum_{p\in\sS} \min_{q\in\sS}\|\hat{\vx}_p^{(n+1)}-\hat{\vx}_q^{(0)}|_2^2+\sum_{q\in\sS} \min_{p\in\sS}\|\hat{\vx}_p^{(n+1)}-\hat{\vx}_q^{(0)}|_2^2,
\end{equation}
where $\sS$ denotes the contact surface between the sensor and the object; $\hat{\vx}=\vx-\bar{\vx}$ for removing the effect of translationa brought by $\vv_r$, and $\bar{\vx}$ denotes the center point of $\vx$.


\begin{figure}[t]
\centering
\includegraphics[scale=0.33]{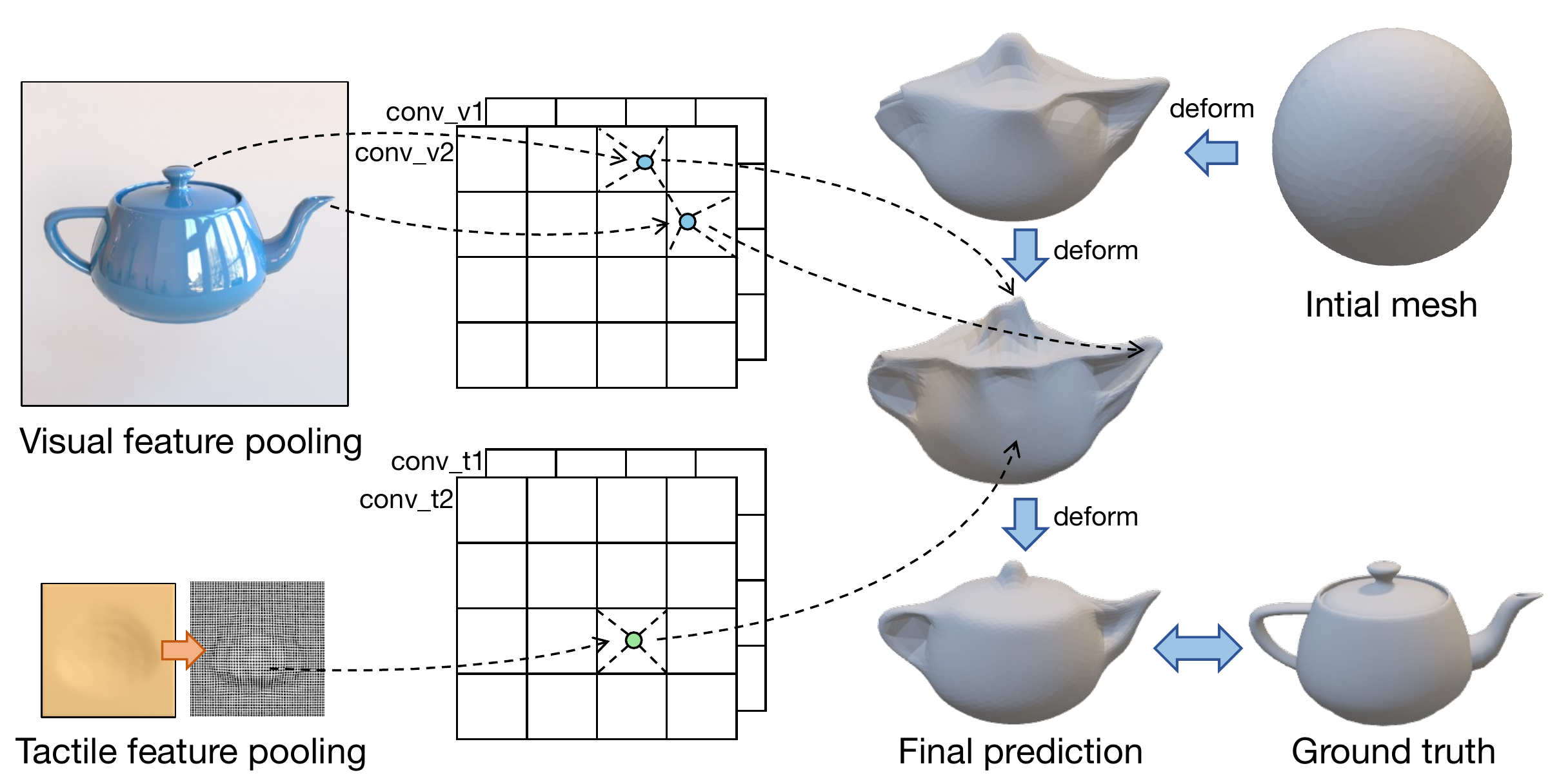}
\caption{The process of the multimodal projection, including visual feature pooling and tactile feature pooling operations. We also show an experimental result of a teapot deforming from the initial sphere to the final prediction.}
\label{pic:projection}
\vskip -0.1 in
\end{figure}

\vspace{-0.08 in}
\subsection{Applications}
\label{sec:application}
\paragraph{Tactile Perception}
Once we obtain the tactile data (namely, the deformation of particles of the sensor), we can apply these data for object recognition. We suppose the tactile deformation of the contact surface as $\mX^{(N)}\in\R^{H\times W\times d}$, where $H$ and $W$ denote the height and the width of the sensor, respectively, and $N$ denotes the final time step. Then, we train a neural network $f$ to predict the object label. Formally,
\vskip -0.1 in
\begin{eqnarray}
\label{eq:perception}
\hat{y} = f(\mX^{(N)}).
\end{eqnarray}
In practice, we prefer to try several attempts of the touch for more accurate recognition. All the deformation outcomes of different touching, denoted as $\{\mX_i^{(N)}\}_{i=1}^I$ will be concatenated along the channel direction, leading to $\mZ\in\R^{W\times H\times (Id)}$, as the input of the network $f$. More implementation details are provided in our experiments.

\begin{figure*}[t]
\centering
\hskip 0.09 in
\includegraphics[scale=0.4]{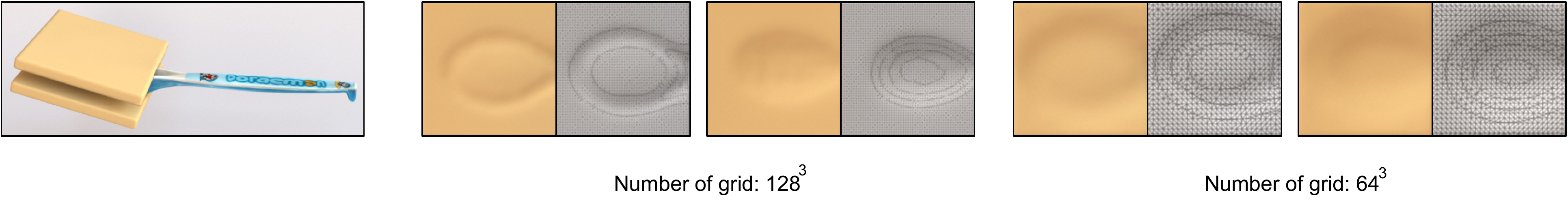}
\small{\caption{Comparison of tactile patterns when pressing a spoon, with different grid number settings. }}
\label{pic:grid}
\end{figure*}

\begin{figure}[t]
\centering
\includegraphics[scale=0.37]{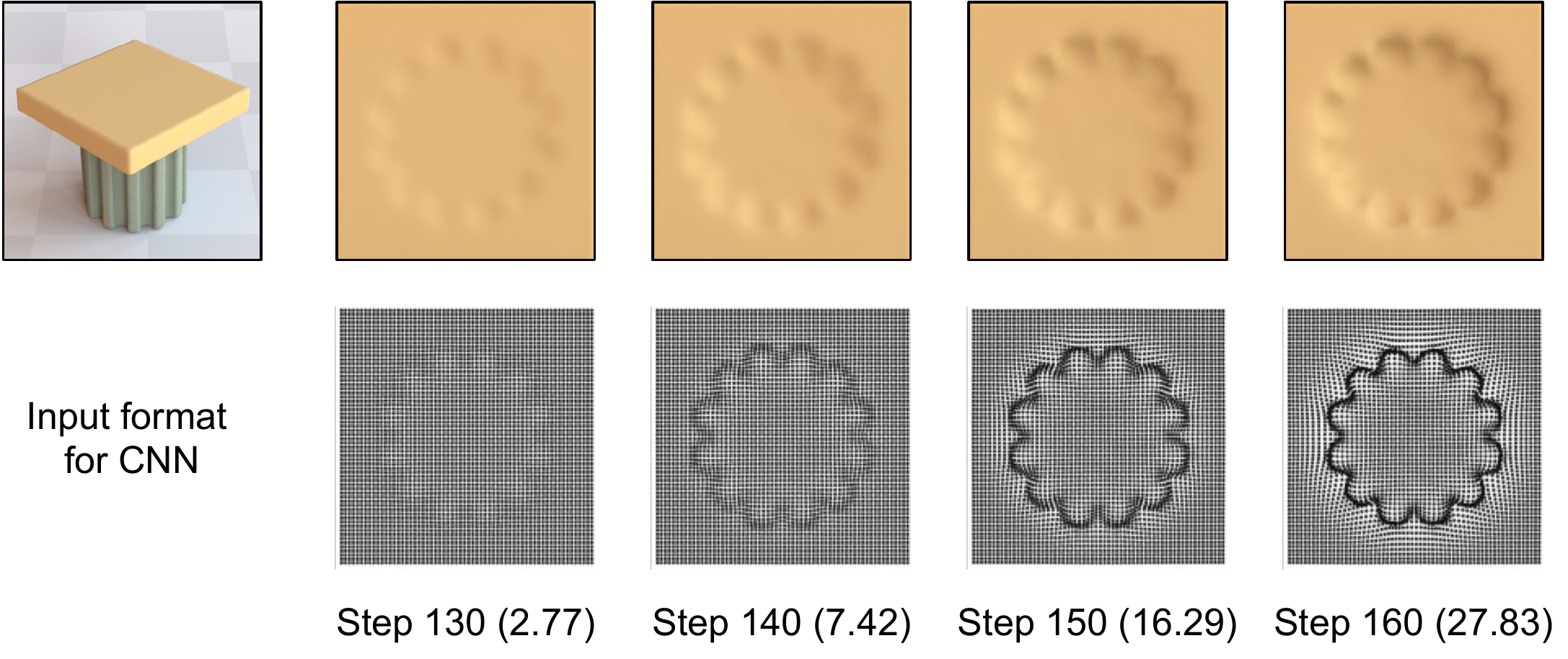}
\caption{Comparison of tactile patterns at different time steps. The extent of deformation increases with the time step increases, followed by the increase of the chamfer distance $l\;(10^{-5})$ as shown in the brackets. We also depict the input format for CNN in the second row.}
\label{pic:extent}
\end{figure}

\begin{figure}[t]
\centering
\includegraphics[scale=0.37]{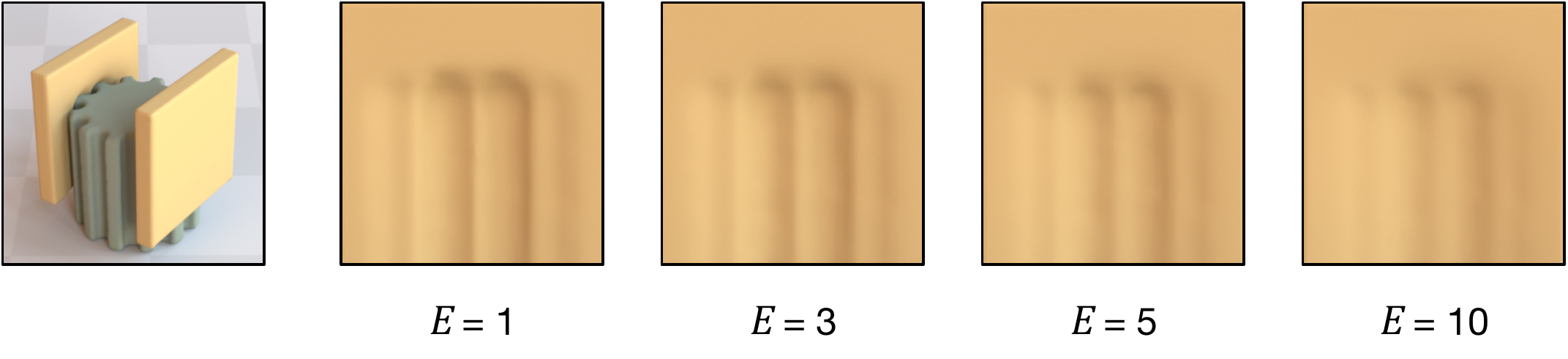}
\caption{Comparison of tactile patterns with different elastic coefficients (Young’s modulus $E$), at the same time step.}
\label{pic:young}
\vspace{-0.5em}
\end{figure}

\paragraph{Tactile-Visual Multimodal Fusion}
Specifically, we make use of the simulated tactile data as complementary information to single-view visual images to predict the 3D geometric model for a manipulated object. During the prediction, the input is composed of an image and a set of tactile data acquired by different grasps of different directions. Here, our backbone is based on the method by~\cite{wang2018pixel2mesh}, where the geometric model is initialized to an ellipsoid mesh and the deformation is realized by a graph neural network (GNN); but different from~\cite{wang2018pixel2mesh}, we further apply the input of tactile data besides the visual images. 

Inspired from~\cite{wang2018pixel2mesh}, the perceptual feature of a certain vertex at a given coordinate is extracted from the pooling of image feature map as well as the pooling of tactile data.  Each vertex on the mesh would be projected to the closest visual/tactile feature, by comparing the distance between the vertex and the tactile center. We adopt zero-padding for each tactile input before the projection, so that the tactile projection process can be unified with the visual projection. The deformation update is described below,
\begin{align}
\vg_v^l&=\text{concat}\left(\vf_v^l,\text{Proj}(\vx_v,\mX_V,\mX_T)\right),\\
\vf_{v}^{l+1}&=\mW_{f0}\vg_v^l+\sum_{v'\in\mathcal{N}(v)}\mW_{f1}\vg_{v'}^l,\\
\vx_{v}^{l+1}&=\mW_{x0}g_v^l+\sum_{v'\in\mathcal{N}(v)}\mW_{x1}\vg_{v'}^l,
\end{align}
where $v$ denotes the index of the vertex on the mesh; $\mathcal{N}(v)$ denotes the neighbors of $v$, which is available as the mesh is initially a sphere; $\vf_v^l,\vx_v^l$ are the feature representation and the learnt coordinate of vertex $v$ for the $l$-th GNN layer, respectively; $\vg_v^l$ is the hidden feature given by the concatenation of $\vf_v^l$ and a multimodal feature projection, as detailed later; $\mW_{f0},\mW_{f1},\mW_{x0},\mW_{x1}$ are learnable weights.
The projection function  $\text{Proj}(\vx_v,\mX_{V},\mX_{T})$ returns the projected coordinates corresponding to the visual map $\mX_{V}$ and the tactile map $\mX_{T}$. 
The projection process is illustrated in Figure \ref{pic:projection}.

\vspace{0.5em}
\section{Experiment}
We implement Algorithm~\ref{alg:tactile_simulation} based on Taichi~\cite{hu2019taichi}. We use Mitsuba~\cite{nimier2019mitsuba} for rendering 3D models. The code source will be made public upon our publication.

\subsection{Effects of Coefficient Settings}
In Figure \ref{pic:grid}, we provide the tactile patterns when pressing a spoon, and compare the patterns under different grid numbers (described in \textsection~\ref{sec:interaction}). We observe that the larger the number of grids is the more fine-grained simulation we will attain. We set the grid number as $128\times128\times128$ considering the trad-off between efficacy and efficiency. To illustrate how the deformation behaves during the whole contact period, in Figure \ref{pic:extent}, we keep pressing the tactile sensor on a gear object and record all results at different time steps. For each time step, we also provide its corresponding format for CNN input, and the Chamfer distance $l$ as described in \textsection~\ref{sec:interaction}; Clearly, the deformation is becoming more remarkable as the contact proceeds.  Figure \ref{pic:young} contrasts the influence by Young's modulus $E$ under the same press displacement. It is shown that the tactile range gets smaller with the increase of the Young's modulus, which is consistent to the conclusion in elastic theory. In our simulation, we choose $E=3$ and $v=0.25$ in Eq.~\ref{eq:phi} as default.

\subsection{Robot Environment Integration}
\begin{figure}[t]
\centering
\includegraphics[scale=0.35]{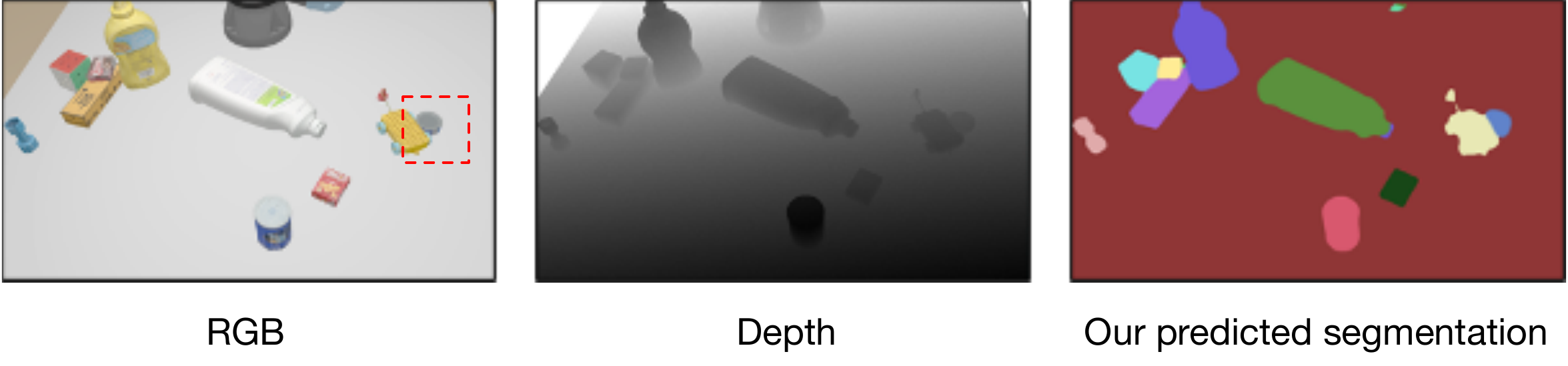}
\caption{Before robot manipulation, we predict the semantic segmentation masks for the objects, based on RGB and depth.}
\label{pic:camera}
\vspace{-1em}
\end{figure}

\begin{figure*}[t]
\centering
\hskip 0.1 in
\includegraphics[scale=0.4]{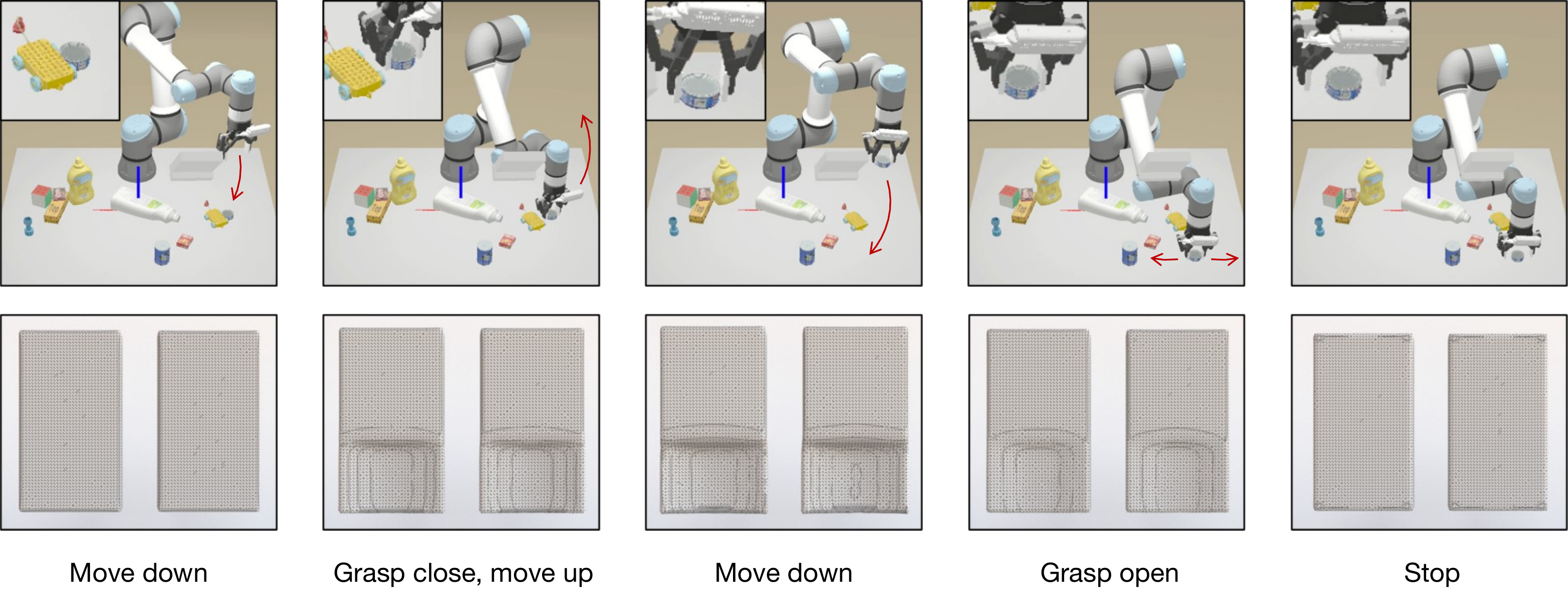}
\vskip -0.05 in
\caption{Illustration of a simulation scene where a robot  picks up a fish can and puts it down. We can see that the corresponding tactile patterns are able to reflect the manipulation process. Besides, when the robot loosens grasping fingers, the tactile sensors also recover to the original state. }
\label{pic:robot-can}
\vskip -0.1 in
\end{figure*}

We integrate our tactile simulation with the robot environment, to perform the pick-and-place task for several different objects. 
As illustrated in Figure \ref{pic:camera}, we first fuse RGB and depth information to get the corresponding semantic segmentation based on the multimodal fusion method in \cite{wang2020cen}. With the segmentation at hand, we detect the 3D position of the can and then pick it up and finally put it down at a different place. The whole process is depicted in Figure \ref{pic:robot-can}, below which we plot the corresponding tactile simulation for each phase. We observe that our tactile simulation does encode the cylinder shape of the can. Besides, the last column  shows that the simulated tactile sensor can return to its original state after the grasping process.

\subsection{Tactile Perception}
\begin{figure}[t]
\centering

\includegraphics[scale=0.212]{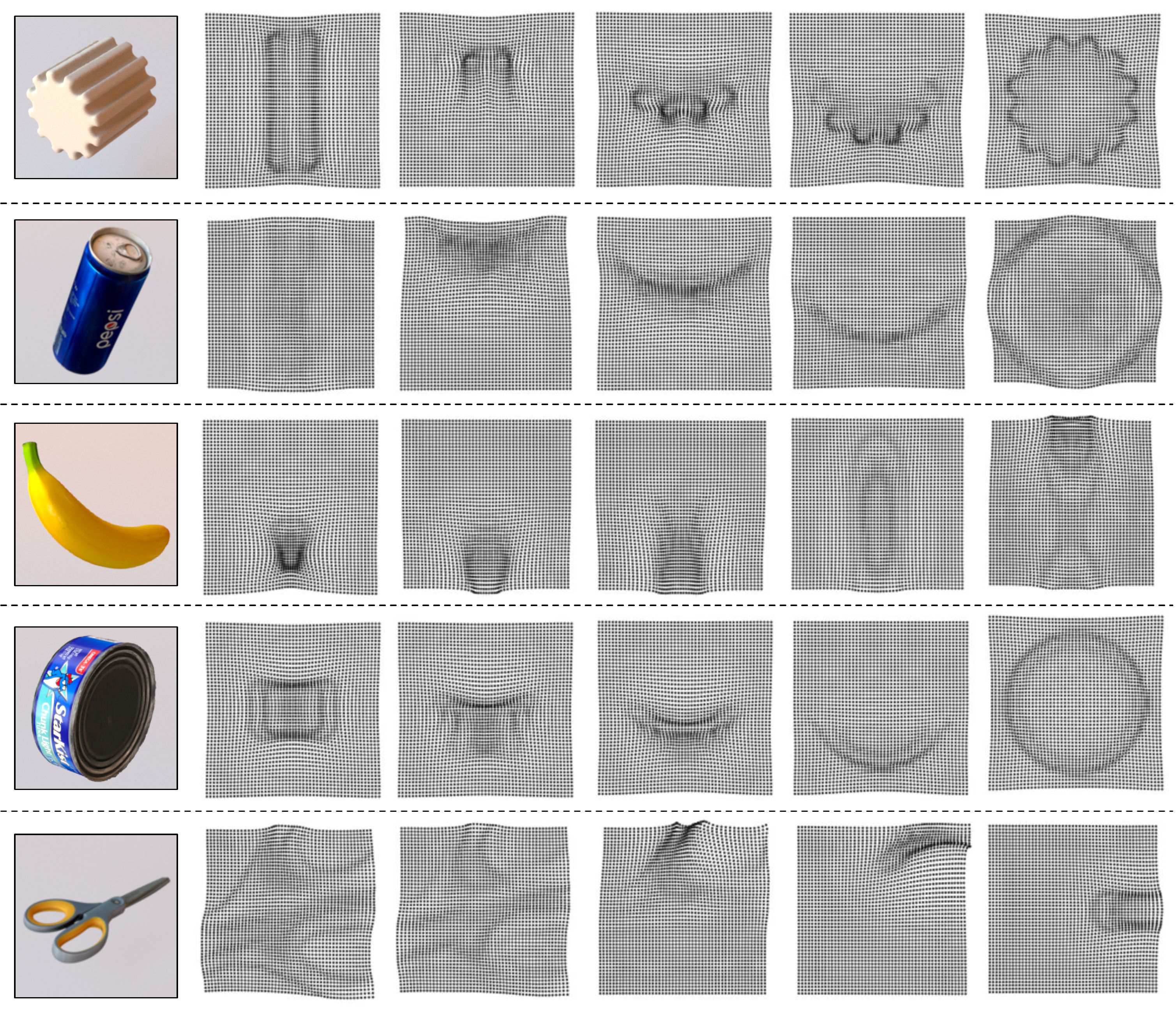}
\vskip -0.05 in
\caption{Visualization of our tactile dataset examples.}
\label{pic:data_example}
\end{figure}

We build a tactile dataset containing 2000/200 tactile images of 10 manipulated objects for training/testing. These patterns are collected through different contact polices including press directions and forces. Figure \ref{pic:data_example} illustrates an example subgroup of the tactile dataset. To predict the object category given tactile patterns, we train a ResNet-32~\cite{he2016deep} from scratch. We also conduct prediction experiments with more than one input tactile image. Specifically, during each training iteration, we randomly choose $N$ tactile images which are yielded by different press direction to the same object. As introduced in~\textsection~\ref{sec:application}, we treat the number of images as the number of input channel. Table \ref{tabs:classification} 
summarizes the accuracies of the tactile perception with the number of inputting images (touches) from 1 to 10. It reads that increasing the number of the touches consistently improves the classification accuracy, and when the number is equal to 10, the accuracy becomes close to $95\%$, which implies the potential usage of our tactile simulation for real robotic perception.

\begin{table}[t]
\centering\caption{Classification accuracies of the tactile perception.}
\vspace{-0.5em}
\resizebox{80mm}{!}{
\begin{tabular}{p{1cm}<{\centering}p{1cm}<{\centering}p{1cm}<{\centering}p{1cm}<{\centering}p{1cm}<{\centering}p{1cm}<{\centering}}
\toprule[0.25mm]
1-touch&2-touch&4-touch&6-touch&8-touch&10-touch  \\
\midrule[0.15mm]
0.375&0.510&0.605&0.810&0.905&0.935\\
\bottomrule[0.25mm]
\end{tabular}}
\label{tabs:classification}
\vskip -0.05 in
\end{table}

\begin{table}[t]
\centering\caption{Results comparison of mesh reconstruction. Evaluation metric: chamfer distance ($\times10^{-3}$).}
\vspace{-0.5em}
\resizebox{80mm}{!}{
\begin{tabular}{p{1cm}<{\centering}|p{1cm}<{\centering}p{1cm}<{\centering}p{1cm}<{\centering}p{1cm}<{\centering}p{1cm}<{\centering}p{1cm}<{\centering}}
\toprule[0.25mm]
\multirow{2}*{Visual}&\multicolumn{5}{c}{Tactile \& Visual}\\

&2-touch  & 4-touch &6-touch&8-touch&10-touch  \\

\midrule[0.15mm]
14.52&10.27&7.71&6.45&5.97&5.23\\
\bottomrule[0.25mm]
\end{tabular}}
\label{tabs:chamfer}
\vskip -0.13 in
\end{table}

\subsection{Tactile-Visual Mesh Prediction}
\begin{figure}[t!]
\centering
\includegraphics[scale=0.23]{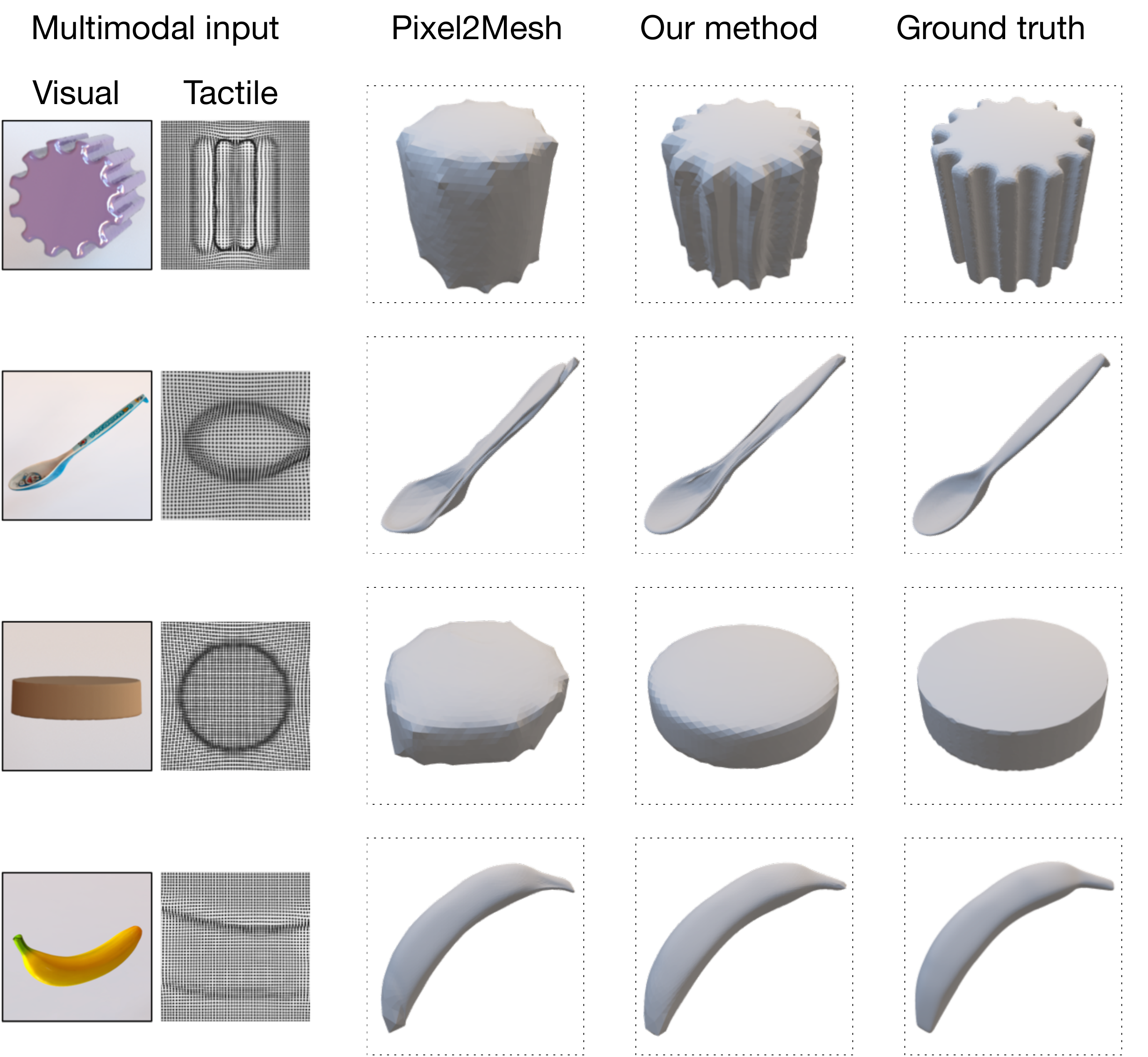}
\caption{Visualization of the mesh reconstruction. Our multimodal method achieves much better performance than using the single-view image only, \emph{i.e.} Pixel2Mesh. Note that we also use other tactile patterns from different directions for reconstructing each mesh apart from the depicted one.}
\label{pic:tactile2mesh}
\vskip -0.1 in
\end{figure}

In this part, we assess the performance of 3D mesh prediction given a single-view image and  a certain number of simulated tactile data by our method. The evaluation is accomplished on 3D mesh models of 20 classes. We collect visual images under random view and tactile data from different press directions. Per each training iteration, we randomly sample 1 visual image and 10 tactile images as the network input. 
Figure \ref{pic:tactile2mesh} displays the generation process of the 3D mesh models. We compare our method with Pixel2Mesh~\cite{wang2018pixel2mesh} that only adopts the visual input. Qualitatively, by adding the tactile input, our approach achieves better prediction output than Pixel2Mesh in Figure~\ref{pic:tactile2mesh}. The quantitative comparison is provided in Table~\ref{tabs:chamfer}, where we calculate the average Chamfer distance as the evaluation metric by sampling both 1000 points from the predicted and target meshes. We vary the input number of tactile images from 2 to 10 in analogy to the perception task before. Even surprisingly, our method with only 2 touches is sufficient to gain smaller reconstruction error than Pixel2Mesh, and the error will become much smaller if we increase the number to 10. The experimental results here well verify the power of our tactile simulation in capturing the fine-grained patterns of the touched object.

\section{Conclusion}
In this work, we propose Elastic Interaction of Particles (EIP), a new method to simulate interactions between the tactile sensor and the object during robot manipulation. Different from existing tactile simulation works, our method is based on the elastic interaction of particles, and allows much more accurate and high-resolution simulation. We integrate the tactile simulation into robotic environment, and conduct two representative groups of experiments to verify the effectiveness of our simulated tactile patterns.


\clearpage
\bibliographystyle{IEEEtran} 
\bibliography{cite}

\end{document}